\documentclass[runningheads]{llncs}

\usepackage{eccv}

\usepackage{eccvabbrv}

\usepackage{graphicx}
\usepackage{booktabs}
\usepackage{amssymb,enumitem}
\usepackage{amsmath}
\usepackage{commath}
\usepackage{multirow,array, bigdelim, makecell, booktabs}
\usepackage{array, caption, floatrow, tabularx}

\usepackage{pifont}
\usepackage{float}
\usepackage{hyperref}
\usepackage{mathrsfs}
\usepackage{dsfont}

\DeclareFloatVCode{beforefloat}{\vspace{4pt}}
\DeclareFloatVCode{afterfloat}{\vspace{-10pt}}

\usepackage[accsupp]{axessibility}  

\usepackage{hyperref}

\usepackage{orcidlink}
\newfloatcommand{capbtabbox}{table}[][\FBwidth]

\DeclareRobustCommand{\etal}{{et al.}}
\DeclareRobustCommand{\eg}{\textit{e.g.}}

\DeclareRobustCommand{\ie}{\textit{i.e.}}

\DeclareRobustCommand{\etal}{{et al.}}

\DeclareRobustCommand{\method}{{CoLeaF}}
\DeclareRobustCommand{\r}{{Reference}}
\DeclareRobustCommand{\a}{{Anchor}}

\newcommand{\linkc}{\textcolor[rgb]{1.0,0.1,0.57}}
\newcommand{\cmark}{\ding{51}}
\newcommand{\xmark}{\ding{55}}

\DeclareRobustCommand{\blue}{\textcolor[rgb]{0.0,0.1,0.95}}

\begin{document}

\title{{\method}: A Contrastive-Collaborative Learning Framework for Weakly Supervised Audio-Visual Video Parsing} 

\titlerunning{{\method}}

\author{Faegheh Sardari\inst{1}\orcidlink{0000-0002-9134-0427} \and
Armin Mustafa\inst{1}\orcidlink{0000-0002-1779-2775} \and
Philip J. B. Jackson\inst{1}\orcidlink{0000-0001-7933-5935} \and \\
Adrian Hilton\inst{1}\orcidlink{0000-0003-4223-238X}}

\authorrunning{F. Sardari et al.}

\institute{Centre for Vision, Speech and Signal Processing (CVSSP), University of Surrey, UK \\
\email{\{f.sardari,armin.mustafa,p.jackson,a.hilton\}@surrey.ac.uk}}

\maketitle

\begin{abstract}
Weakly supervised audio-visual video parsing (AVVP) methods aim to detect audible-only, visible-only, and audible-visible events using only video-level labels. Existing approaches tackle this by leveraging unimodal and cross-modal contexts. However, we argue that while cross-modal learning is beneficial for detecting audible-visible events, in the weakly supervised scenario, it negatively impacts unaligned audible or visible events by introducing irrelevant modality information. In this paper, we propose {\method}, a novel learning framework that optimizes the integration of cross-modal context in the embedding space such that the network explicitly learns to combine cross-modal information for audible-visible events while filtering them out for unaligned events. Additionally, as videos often involve complex class relationships, modelling them improves performance. However, this introduces extra computational costs into the network. Our framework is designed to leverage cross-class relationships during training without incurring additional computations at inference. Furthermore, we propose new metrics to better evaluate a method's capabilities in performing AVVP. Our extensive experiments demonstrate that {\method} significantly improves the state-of-the-art results by an average of 1.9\% and 2.4\% F-score on the LLP and UnAV-100 datasets, respectively. Code is available at: \href{https://github.com/faeghehsardari/coleaf}{\linkc{https://github.com/faeghehsardari/coleaf}}.
\end{abstract}

\section{Introduction}
\label{sec:intro}
\begin{figure}[t]
\CenterFloatBoxes
\begin{floatrow}
\ffigbox
  {
  \scalebox{1.0}{
  \includegraphics[width=1.0\linewidth]{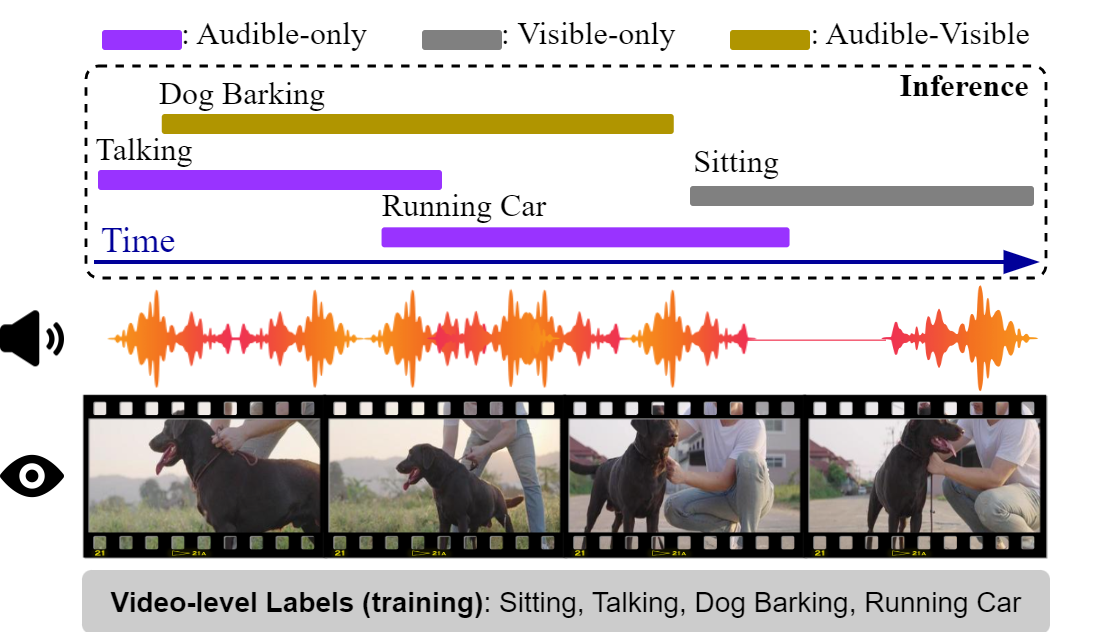}
  \caption{An example of weakly supervised AVVP task. The method learns to temporally detect audible-only, visible-only, and audible-visible events in a video, using only video-level labels during training.}
  \label{fig:avvp}}}
\killfloatstyle
\ttabbox
  {\caption{Performance of CMPAE \cite{gao2023collecting} when using only unimodal (U) information versus exploiting both unimodal and cross-modal contexts (U+C) in detecting audible-only (A), visible-only (V) and audible-visible (AV) event types, in terms of true positive (TP), true negative (TN), false positive (FP), and false negative (FN) rates.\vspace{-3mm}}
  \scalebox{0.8}
{
  \setlength{\tabcolsep}{1.0pt} 
    \renewcommand{\arraystretch}{1.0} 
  \begin{tabular}{c c c c c c c } \specialrule{.2em}{.1em}{.1em}
   \multirow{2}{*}{\bf Metric} &~\phantom & \bf A &~~\phantom & \bf V &~~\phantom & \bf AV \cr \cmidrule{3-3} \cmidrule{5-5} \cmidrule{7-7} 
   &~\phantom & \bf U/(U+C) &\phantom & \bf U/(U+C)\bf  &\phantom & \bf U/(U+C) \cr \specialrule{.2em}{.1em}{.1em}
   \bf TP & \phantom &{\bf{43.5}}~/~42.7 & \phantom &{\bf{67.8}}~/~66.4 & \phantom &80.8~/~{\bf {84.6}}\cr 
   \bf TN &\phantom & {\bf{97.6}}~/~97.4&\phantom & {\bf{97.3}}~/~97.0&\phantom & 95.2~/~{\bf{95.6}}\cr
   \bf FP & \phantom &{\bf{0.4}}~/~0.6&\phantom & 0.9~/~{\bf{0.7}}& \phantom &{\bf{3.3}}~/~3.8\cr
   \bf FN &\phantom & {\bf{94.7}}~/~95.5&\phantom & {\bf{97.7}}~/~99.2&\phantom & 33.2~/ {\bf{29.9}}\cr\specialrule{.2em}{.1em}{.1em}
  \end{tabular}}}
  {
  \label{tab:different modality}}
\end{floatrow}
\end{figure}

Audio-visual learning has emerged as a crucial field within computer vision, enabling AI systems to develop a more nuanced understanding of the world. Most audio-visual tasks (\eg, audio-visual event localization \cite{tian2018audio, Xia_2022_CVPR,geng2023dense} and audio-visual question answering \cite{Yun_2021_ICCV,li2022learning,nadeem2023cad}) assume that audio and visual data are temporally aligned and share a common concept. However, in practice, this assumption is not always valid. For instance, a baby might cry out of the camera's view. The Audio-Visual Video Parsing (AVVP) task \cite{tian2020unified} instead introduces an unaligned setting of audio-visual learning to recognize and localize audible-only, visible-only, and audible-visible events in a video. Since training a fully supervised AVVP model requires modality-specific densely annotated data, making it a costly endeavour, the AVVP task is tackled in a weakly supervised scenario where only video-level labels are available, see Fig.\ref{fig:avvp}. 
To address the weakly supervised AVVP task, all existing approaches, such as \cite{tian2020unified, yu2022mm, gao2023collecting}, incorporate unimodal and cross-modal contexts. However, we argue that while extracting cross-modal information is beneficial for multi-modal learning tasks, in the weakly supervised scenario, it can negatively impact the detection of audible-only and visible-only events (i.e., unaligned audible-visible events) by introducing irrelevant modality information. We investigated this in Table \ref{tab:different modality} by comparing the performance of CMPAE \cite{gao2023collecting} with and without using cross-modal contexts. The results show that leveraging cross-modal contexts improves the performance on average in detecting audible-visible events, while in detecting unaligned events, a network focusing solely on unimodal information yields better results. These findings motivated us to raise this question: {\bf{ Can we optimize the integration of cross-modal context in the embedding space such that the network explicitly learns to leverage cross-modal information for audible-visible events, while filtering them out for unaligned events?}} To achieve this, we introduce {\method}, a novel learning framework for the weakly supervised AVVP task. {\method} relies on two network branches: {\r} and {\a}. {\r} focuses on unimodal information, while {\a} incorporates unimodal and cross-modal contexts. Both branches process the same audio and visual input signals and are trained simultaneously using the video-level labels. Additionally, during training, {\r} is leveraged by a novel contrastive objective to optimize cross-modal learning in {\a} based on the event types present in the input video. The proposed objective encourages {\a} to learn embedding representations similar to those learned in {\r}, but the strength of the encouragement adapts to the degree of unaligned audible-visible events present in the input video. More unaligned events lead to a stronger encouragement, while fewer unaligned events lead to a weaker encouragement. The unalignment degree is computed from the {\r}’s output predictions. Since the optimization process relies on the performance of {\r}, its effectiveness may diminish if {\r} is solely trained on `modality-agnostic' video-level labels. To address this without requiring additional annotations, {\a} collaboratively distils free modality-aware pseudo-labels for {\r}. After training, {\a} is deployed for inference.

To effectively perform the AVVP task, we should exploit unimodal and cross-modal contexts while aware of class dependencies. For example, this is often seen when musical instruments are played alongside people singing. Authors in \cite{lin2021exploring, mo2022multi} design networks to explicitly learn cross-class relationships. However, their approaches are computationally expensive. For example, in \cite{lin2021exploring}, to capture cross-class dependencies, they extend their model to map each temporal segment of the video into the maximum number of event classes, and apply attention layers within the event classes for each segment individually. For a video with a length of T segments and C class categories, this led to a significant computational overhead with a ratio of $T\times C\times C$. This key challenge motivated us to raise this novel question: {\bf Can we take  advantage of explicitly modelling co-occurrence class relationships without imposing the overhead of heavy computational costs?} To achieve this, in {\method}, {\r} that is employed only during training is designed to explicitly learn co-occurrence class relationships and distil this knowledge to {\a}. This novel design allows us to benefit from cross-class relationships during training without introducing computational overhead on {\a} that is employed at inference. 

For a deeper understanding of how our proposed framework models unaligned events, we introduce novel metrics for AVVP. Current metrics evaluate the method’s performance in detecting audible-only and visible-only events by comparing the estimated audio and visual event proposals individually with audio and visual ground-truth labels. In our proposed metrics, we consider both audio and visual modalities, as comparing them individually can lead to the mistaken interpretation of an audible-visible event as both audible-only and visible-only. Our key contributions are summarized as: (i) we pioneer the first exploration in explicitly optimizing the integration of cross-modal context for the weakly supervised AVVP task, (ii) we propose an approach that allows the network to explicitly learn cross-class relationships without adding computational overhead during inference. This can benefit various computer vision tasks that demand a deeper understanding of interactions between classes, (iii) we introduce novel metrics for effective evaluation of AVVP which are beneficial for future development of this task, (iv) we evaluate {\method} on two challenging publicly available datasets and improve the state-of-the-art results on average by {1.9\% and 2.4\% F-score on LLP and UnAV-100 datasets, respectively}, and (v) we perform extensive ablation studies to evaluate our proposed framework’s design.

\section{Related Works}
\label{sec: related works}
Tian~\etal~\cite{tian2020unified} defined the AVVP task and also introduced the LLP dataset for this purpose.  Due to the laborious nature of the labelling process for this task, for the training set, they obtained only video-level labels, so the learning process is performed in a weakly supervised setting. To perform the task, they design the Hybrid Attention Network (HAN), where they extract and combine unimodal and cross-modal contexts from audio and visual data by applying self-attention and cross-attention mechanisms and introduce Multimodal Multiple Instance Learning (MMIL) to facilitate network training under weakly supervised setting. Following HAN, the proposed approaches can be divided into the two categories:

\noindent{\bf Architectural Improvements -- } To better explore unimodal and cross-modal contexts, the authors in \cite{yu2022mm, jiang2022dhhn} expand HAN to a multi-scale pyramid network. Alternatively, Gao~\etal~\cite{gao2023collecting} attempt to decrease the uncertainty of HAN’s predictions under weakly supervised learning by introducing a modality presence-absence confidence term in the cross-entropy loss function based on Subjective Logic Theory \cite{sensoy2018evidential}. Similar to the network architecture, their loss confidence term is also explicitly derived from the integration of unimodal and cross-modal information. However, as stated in Section \ref{sec:intro}, the cross-modal learning in these architectures negatively affects the detection of unaligned events. In contrast, in our proposed framework, the network explicitly learns to filter out cross-modal contexts for unaligned events. On the other hand, \cite{lin2021exploring, mo2022multi} additionally learn cross-class dependencies. Although they improve HAN’s performance, they suffer from computational complexity. In contrast, our framework leverages the class relationships during training without incurring additional costs in inference.

\noindent{\bf Label Refinement -- } The core goal of these approaches \cite{wu2021exploring, jiang2022dhhn, cheng2022joint, gao2023collecting, zhou2023improving} is to enhance the performance of existing network architectures by obtaining modality-specific knowledge for training. For example, the authors in \cite{gao2023collecting} devised a two-stage approach in which they initially extract offline modality-specific annotations by applying large-scale pre-trained models such as CLAP \cite{wu2023large} and CLIP \cite{radford2021learning} on the audio and visual data. Subsequently, they train a network backbone from scratch using the additional labels and a cross-entropy loss function. However, the training process of such approaches is complicated and inefficient. Alternatively, \cite{jiang2022dhhn} propose an online method where the estimated probabilities of a network, focusing on individual modalities, guide the final estimation of the AVVP task through the mean square error loss. These approaches may optimize the integration of cross-modal information as they obtain hard/soft modality-aware labels for training. However, their optimization process is implicit and suboptimal. In contrast, our proposed approach explicitly optimizes cross-modal learning in the embedding space. Furthermore, in these approaches, the pseudo labels are obtained from {`modality-agnostic'} video-level labels, making them unreliable. However, our online collaborative solution mitigates the impact of weak labels on our optimization process. 

\section{Methodology}
In this section, we first define the weakly supervised audio-visual video parsing task, followed by a brief review of the HAN network \cite{tian2020unified}. Then, we elaborate on our proposed framework {\method}. 
\subsection{Preliminaries}
\noindent{\bf Problem Formulation -- } In weakly supervised AVVP, our goal is to detect audible-only, visible-only, and audible-visible events in a given video by training from video-level labels, as defined in \cite{tian2020unified}. For the video sequence $S=\{A_t, V_t\}^{T}_{t=1}$ with a length of T segments, $A_t$ and $V_t$ represent the audio and visual signals at the $t^{th}$ segment, respectively. At the inference, the network task is to provide the multi-class event labels $Y_t =\{y^{a}_{t,i}, y^{v}_{t,i}, y^{av}_{t,i}\}^{C}_{i=1}$ for each time segment of $S$, where $y^{a}_{t,i}, y^{v}_{t,i}, y^{av}_{t,i} \in\{0, 1\}$ indicate audible-only, visible-only, and audible-visible events, respectively, and $C$ represents the maximum number of event classes in the dataset. However, during training, we lack access to the {`modality-specific segment-level'} labels $Y_t$, having only the {`video-level'} labels that are {`modality-agnostic'} ${Y} =\{{y}_{i}\}^{C}_{i=1}$, where $y_i\in\{0, 1\}$. 

\noindent{\bf Hybrid Attention Network (HAN) -- } HAN \cite{tian2020unified} is a popular baseline for the weakly supervised AVVP task. In HAN, first, two pre-trained models are applied on the audio and visual signals of the input video $S$ to extract segment-level audio and visual input tokens, $F^a=\{f^{a}_{t}\}^{T}_{t=1}$ and $F^v=\{f^{v}_{t}\}^{T}_{t=1}$, respectively. Then, it aggregates unimodal and cross-modal information through applying and combining self-attention and cross-attention mechanisms \cite{vaswani2017attention} on the input tokens 
\begin{equation}
{\hat{f}}^{a}_{t} = {f}^{a}_{t} + \text{Self-attn}({f}^{a}_{t}, {F}^{a}) + \text{Cross-attn}({f}^{a}_{t}, {F}^{v}),
\end{equation}\vspace{-4mm}
\begin{equation}
{\hat{f}}^{v}_{t} = {f}^{v}_{t} + \text{Self-attn}({f}^{v}_{t}, {F}^{v}) + \text{Cross-attn}({f}^{v}_{t}, {F}^{a}),
\end{equation}
\noindent Finally, HAN uses a share fully connected layer FC to predict multi-class segment-level audio and visual event probabilities from the audio and visual output tokens as ${P}^{\phi} = Sigmoid(FC(\hat{F}^{\phi}))$, where $P^{\phi}=\{{p}^{\phi}_{t}\in \rm {I\!R}^{1\times C}\}^{T}_{t=1}$ and $\phi\in\{a,v\}$. During the inference $P^{\phi}$ is utilized to predict segment-level audible-only, visible-only, and audible-visible events. However, during the training, as we only have video-level labels, HAN applies a Multimodal Multiple Instance Learning (MMIL) module to provide video-level probabilities as 
\begin{equation}
\small
\label{eq: pt1}
W^{\vartriangle} = \textit{Soft-max}_{t}(FC_{t}([\hat{F}^a;\hat{F}^v]), ~
W^{\blacktriangle} = \textit{Soft-max}_{av}(FC_{av}([\hat{F}^a;\hat{F}^v]), 
\end{equation}\vspace{-4mm}
\begin{equation}
\small
\label{eq: pt2}
\mathbb{P} = \sum_{t=1}^{T} \sum_{m=1}^{M} (W^{\vartriangle}_{t,m}\odot W^{\blacktriangle}_{t,m}\odot{P}_{t,m}), ~ \text{where } {P} = [P^a; P^v], 
\end{equation}
and $M\hspace{-0.4em}=\hspace{-0.4em}2$ refers to audio and visual modalities, ${P},W^{\vartriangle},W^{\blacktriangle}\in \rm {I\!R}^{T\times 2\times C}$, $\mathbb{P}\in \rm {I\!R}^{1\times C}$, and $[;]$ and $\odot$ indicates the concatenation and Hadamard product operations, respectively.  $\textit{Soft-max}_{t}$ and $\textit{Soft-max}_{av}$ represent the Softmax layers along the temporal and audio-visual dimensions, respectively. With the video-level probabilities $\mathbb{P}$, and the video-level labels $Y$, HAN is optimized through Binary Cross Entropy (BCE) loss $\mathcal{{L}}_{video} = BCE(Y, \mathbb{P})$. 

\begin{figure}[t]
  \centering
  \includegraphics[width=1.0\linewidth]{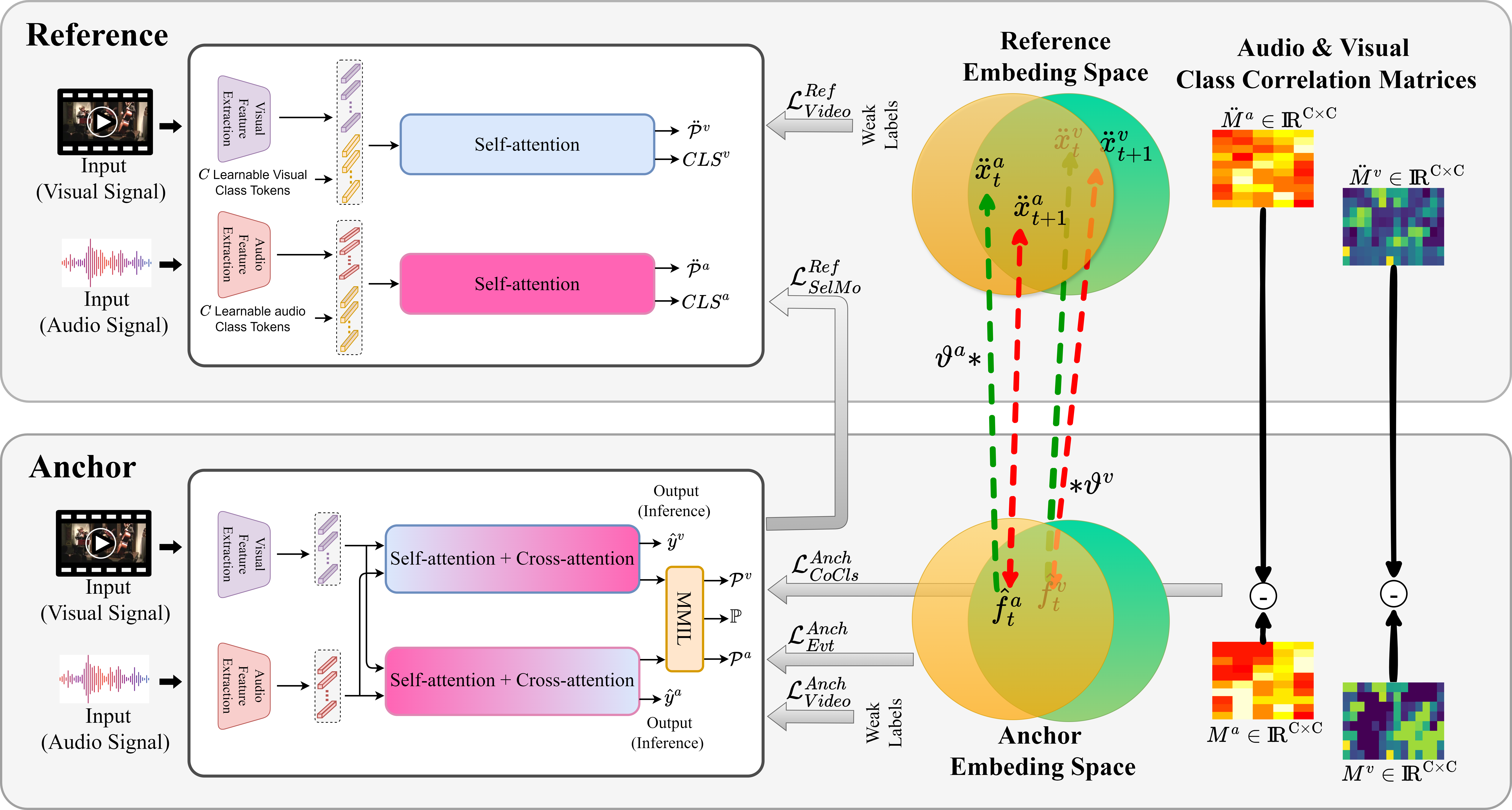}
  \caption{The overall scheme of {\method} for the weakly supervised AVVP task. {\method} has two network branches, {\r} and {\a}. The {\r} branch focuses on unimodal information and cross-class relationships, while the {\a} branch exploits both unimodal and cross-modal contexts. The branches are trained simultaneously through the conventional video-level losses $\{\mathcal{{L}}^{Ref}_{video},\mathcal{{L}}^{Anchr}_{video}\}$ and our novel contrastive and knowledge distillation losses {$\{\mathcal{L}^{Anch}_{Evt}, \mathcal{L}^{Ref}_{SelfMo},\mathcal{L}^{Anchr}_{CoCls}\}$}. In inference, {\a} is deployed for the AVVP task.}
  \label{fig:2CoLeaF}
  \vspace{-3mm}
\end{figure}

\subsection{{\method}: Contrastive-Collaborative Learning Framework}
In this section, we introduce {\method}, a novel learning framework designed to address two key challenges in the weakly supervised AVVP task: (i) effectively integrating the cross-modal contexts with the lack of modality-specific annotations, and (ii) efficiently learning cross-class relationships. {\method} relies on two network branches, {\r} and {\a}, as shown in Fig. \ref{fig:2CoLeaF}. The {\r} branch, used only during training, is dedicated to exploiting unimodal and co-occurrence class relationships. The {\a} branch exploits unimodal and cross-modal contexts and serves as the main network branch deployed during inference. The branches receive the same audio and visual input data and are trained simultaneously using two types of losses: (a) the conventional video-level weakly supervised loss, as in HAN, and (b) our proposed collaborative losses, including event-aware contrastive loss, co-occurrence class knowledge distillation loss, and self-modality-aware knowledge distillation loss. In the following, we explain the details of each branch and the losses.
\noindent{\bf {\r} Branch -- } This branch addresses the weakly supervised AVVP task by primarily focusing on unimodal information. This can be achieved by applying two self-attention layers on the audio and visual input tokens individually. However, as videos with multi-label events contain complex relationships between the event classes, we simultaneously leverage this information to improve the method’s performance. To model explicitly the cross-class relationships, inspired by \cite{mo2022multi}, we first concatenate the audio and visual input tokens, $F^{a}\hspace{-0.3em}=\hspace{-0.3em}\{f^{a}_t\}^{T}_{t=1}$ and $F^{v}\hspace{-0.3em}=\hspace{-0.3em}\{f^{v}_t\}^{T}_{t=1}$, with learnable audio and visual class tokens, $C^{a}\hspace{-0.3em}=\hspace{-0.3em}\{c^{a}_i\}^{C}_{i=1}$ and $C^{v}\hspace{-0.3em}=\hspace{-0.3em}\{c^{v}_i\}^{C}_{i=1}$, 
\begin{equation}
    {X}^{\phi}=\{{x}^{\phi}_{n}\}^{T + C}_{n=1}=[\{f^{\phi}_t\}^{T}_{t=1};\{c^{\phi}_i\}^{C}_{i=1}],
\end{equation}
where $\phi\in\{a,v\}$ and $[;]$ indicates the concatenation operations. We then forward them into our self-attention backbones to exploit both intra-modality and class relationships,
\begin{equation}
\{\ddot{x}^{\phi}\}^{T+C}_{n=1}= \{\text{Self-attn}_{\phi}({x}^{\phi}_{n}, {X}^{\phi})\}^{T + C}_{n=1}.
\end{equation}
We take the outputs of self-attention layers which correspond to the audio and visual input tokens, $\ddot{F}^{\phi}=\{\ddot{x}^{\phi}_n\}^T_{n=1}$, to predict segment-level audio and visual event probabilities through a fully connected layer as $\{\ddot{p}^{\phi}_{t}\}^T_{t=1} = Sigmoid(FC_{\phi}(\ddot{F}^{\phi}))$, where $\ddot{p}^{\phi}_{t}\in \rm {I\!R}^{1\times C}$. Subsequently, the audio and visual event probabilities are computed as ${\ddot{\mathcal{P}}}^{\phi} =\sum_{t=1}^{T}(\ddot{w}^{\phi}_{t}\odot \ddot{p}^{\phi}_{t})$
\begin{equation}
\small
    {\ddot{\mathcal{P}}}^{\phi} =\sum_{t=1}^{T}(\ddot{w}^{\phi}_{t}\odot \ddot{p}^{\phi}_{t}),
\end{equation}
where ${\ddot{\mathcal{P}}}^{\phi},\ddot{w}_{t}\in \rm {I\!R}^{1\times C}$, and $\ddot{w}_{t}$ are temporal learnable weight. With the audio and visual  event probabilities and video-level annotations $Y\in \rm {I\!R}^{1\times C}$, the {\r} branch is optimized through the video-level loss $\mathcal{{L}}^{Ref}_{video} = \sum_{\phi\in\{a,v\}}BCE(Y, \ddot{\mathcal{P}}^{\phi})$.

Optimizing the {\r} branch by relying only on the $\ddot{F}^{\phi}$ output tokens does not guarantee that the input class tokens learn to model the class events. To ensure that they represent the classes effectively, we also utilize the output tokens of self-attention layers which correspond to the input class tokens, $\{\ddot{c}^{\phi}\}^{C}_{i=1}=\{\ddot{x}^{\phi}_n\}^{T+C}_{n=T+1}$, for the network optimization. To achieve this, we apply a global average pooling layer on these tokens along their embedding dimension to convert them into a 1-D dimensional vector, followed by a sigmoid activation function to convert them into range 0 and 1 as $CLS^{\phi}=\{ Sigmoid(AvgPool(\ddot{c}^{\phi}_i))\}^{C}_{i=1}$. Then, they are supervised by video-level annotations and cross entropy loss. Therefore, the video-level loss is extended as $\mathcal{{L}}^{Ref}_{video} = \sum_{\phi\in\{a,v\}}\hspace{-0.3em}BCE(Y, \ddot{\mathcal{P}}^{\phi}) + BCE(Y, {CLS}^{\phi})$.

\noindent{\bf {\a} Branch -- } This  branch tackles the weakly supervised AVVP task by exploiting both unimodal and cross-modal contexts. To achieve this, a backbone that performs the AVVP task using self-attention and cross-attention mechanisms (\eg, HAN) can serve as {\a} to predict segment-level and video-level probabilities $P^{\phi}=\{p^{\phi}_t\in \rm {I\!R}^{1\times C}\}^T_{t=1}$ and $\mathbb{P}\in \rm {I\!R}^{1\times C}$ (Eqs \ref{eq: pt1}-\ref{eq: pt2}), respectively, from the audio and visual input tokens $F^{\phi}$. The segment-level labels are employed for inference phase, as well as later in this section for {our proposed self-modality-aware knowledge distillation loss}, while the video-level probabilities are utilized to compute the video-level loss $\mathcal{{L}}^{Anch}_{video}=BCE(Y,\mathbb{P})$, as in HAN network.

\noindent {\bf Event-Aware Contrastive Learning --} We argue that while leveraging cross-modal information is beneficial in the detection of audible-visible events, in the weakly supervised scenario, it can negatively impact the detection of audible-only and visible-only events by introducing irrelevant modality information. Our findings in Table \ref{tab:different modality} support this argument. To tackle this challenge, we introduce contrastive learning to encourage the learned representations in the {\a} branch to become similar to those of the {\r} branch for unaligned audio-visible event types. To achieve this, for each audio and visual output tokens of the {\a} branch ${\hat{f}}^{\phi}_{t}$, the output tokens of the corresponding segment from the {\r} branch $\ddot{x}^{\phi}_t$ can be chosen as positive samples, and the output tokens of the rest of its segments can be considered as negative set. Subsequently, the {\a} branch can be optimized through Noise Contrastive Estimation (NCE) loss \cite{gutmann2010noise}.  In this formulation, regardless of the event types in a video, {\a} is encouraged to align with {\r}. However, a video can contain multiple events spanning audible-only, visible-only, and audible-visible, and our goal is to get close to the tokens' representations of the {\r} branch for only unaligned audio-visible event types. We address this issue by introducing a novel event-aware NCE loss that adjusts the strength of contrastive learning encouragement based on the degree of audible-only and visible-only events present in the input video

\begin{equation}
\small
    \mathcal{L}^{Anch}_{Evt} = - \frac{1}{T} * \sum_{\phi\in\{a,v\}}\vartheta^{\phi} *\sum_{t=1}^{T}\log{\frac{\exp{({{\hat{f}}^{\phi^\intercal}_{t}}.{\ddot{x}^{\phi}}_{t}/\tau)}}{\sum_{n=1,n\neq t}^{T}\exp{({{\hat{f}}^{\phi^\intercal}_{t}}.{\ddot{x}^{\phi}}_{n}/\tau)}}},
\end{equation}

\noindent where weight $\vartheta^{\phi}$ is computed from a set of pseudo-labels $\ddot{G}^{\phi}=\{\ddot{g}^{\phi}_i\}^C_{i=1}$ distilled from the {\r} branch 
\begin{gather}
    \small
    \ddot{g}^{\phi}_i = \begin{cases}
                           1 & \text{if $\ddot{\mathcal{P}}^{\phi}_i > \theta$}\\                                           0 & \text{else}
             \end{cases},\\
    N^{a} = \sum_{c=1}^{C} \ddot{g}^{a}_i \odot (1-\ddot{g}^{v}_i), N^{v} = \sum_{c=1}^{C} \ddot{g}^{v}_i \odot (1-\ddot{g}^{a}_i), N^{av} = \sum_{c=1}^{C} \ddot{g}^{a}_i \odot \ddot{g}^{v}_i,\\
    \vartheta^{a} = \frac{N^a}{N^a + N^{av}}, \text{   and    } \vartheta^{v} = \frac{N^v}{N^v + N^{av}}.
\end{gather}
$N^a$, $N^v$, and $N^{av}$ indicate the number of audible-only, visible-only, and audible-visible events in the video, respectively. $\vartheta^{\phi}$ reflects the degree of encouragement in the contrastive learning and takes a value between 0 and 1 where 0 indicates there is no event or all the events are audible-visible, resulting in no encouragement, and 1 indicates all the events are unaligned, resulting in maximum encouragement.

\vspace{1.0mm}
\noindent {\bf Self-Modality-Aware Knowledge Distillation -- }
If the {\r} branch relies solely on {`modality-agnostic'} video-level labels for training, we face two key limitations: (i) {\r} cannot acquire rich representations necessary to teach {\a}, and (ii) the proposed event-aware contrastive loss may not effectively optimize cross-modal learning since it relies on the predicted event types by {\r}. To overcome these \textit{without additional annotations}, we propose to leverage \textit{free modality-aware knowledge} from the {\a} branch to improve the performance of the {\r} branch. To perform this, we first distil pseudo labels ${G}^{\phi}=\{{g}^{\phi}_i\}^C_{i=1}$ from the estimated audio and visual event probabilities in the {\a} branch,
\begin{equation}
{g}^{\phi}_i = \begin{cases}
                           1 & \text{if ${\mathcal{P}}^{\phi}_i > \theta$}\\                                           0 & \text{else}
             \end{cases}
\end{equation} 
Then, ${G}^{\phi}$ are used to compute our proposed {self-modality-aware knowledge distillation loss} for the {\r} branch as $\mathcal{{L}}^{Ref}_{SelfMo}=\sum_{\phi\in\{a,v\}}BCE({G}^{\phi}, \mathcal{\ddot{P}}^{\phi})$.

\vspace{1mm}
\noindent{\bf{Co-occurrence Class Knowledge Distillation -- }} The primary goal of modeling the co-occurrence class relationships explicitly in the {\r} branch is to transfer them to the {\a} branch so that we can benefit from this knowledge during inference without suffering from the associated computation overhead. Optimizing the {\a} branch through our proposed contrastive learning may implicitly transfer class relationships from the {\r} branch to the {\a} branch, but cannot fully transfer them because it is not designed for this purpose. To explicitly transfer the cross-class relationships, we introduce the co-occurrence class knowledge distillation loss, where we first obtain the audio and visual class correlation matrices, ${\ddot{M}^{\phi}}\in \rm{I\!R}^{C\times C}$ and ${{M}^{\phi}}\in \rm{I\!R}^{C\times C}$, for the {\r} and {\a} branches as 
\begin{equation}
\small
{\ddot{M}^{\phi}}_{i, j} = \ddot{\mathcal{P}}^{\phi}_{i}*\ddot{\mathcal{P}}^{\phi}_{j} \text{, and }{{M}^{\phi}}_{i,j} = {\mathcal{P}}^{\phi}_{i}*{\mathcal{P}}^{\phi}_{j}.
\end{equation}
By computing the correlations, we quantify the relationships between each pair of classes. Then, we encourage the {\a} branch to mimic the audio and visual cross-class correlations of the {\r} branch through the mean square error (MSE) as 
$\mathcal{{L}}^{Anch}_{CoCls} = \sum_{\phi\in\{a,v\}}MSE(\ddot{M}^{\phi},{M}^{\phi})$.


\section{New Metrics for AVVP}
\label{sec: metrics}
Current approaches utilize 5 metrics, under segment-level and event-level, to evaluate the method's performance for the AVVP task. These metrics include F-score for audible-only (A), visible-only (V), and audible-visible (AV) events separately, as well as the average F-score of the A, V, and AV (Type@AV), and the overall F-score for all A and V events within each sample (Event@AV). To compute these metrics, they first employ a threshold on the estimated audio and visual event probabilities to identify audible and visible events, $\hat{y}^a_t$ and $\hat{y}^v_t$, respectively. Then, the F-score for A and V are computed by comparing $\hat{y}^a_t$ and $\hat{y}^v_t$ with their corresponding audio and visual ground-truth labels, ${y}^a_t$ and ${y}^v_t$, respectively. The F-score for AV instead is computed by simultaneously considering both audible and visible events and their ground-truth labels.

We argue that the A and V metrics fail to accurately evaluate the method's performance in detecting audible-only and visible-only events because they consider only their corresponding modalities. In contrast, we believe that both modalities should be considered when computing A and V, similar to AV.  For instance, consider a scenario where the network predicts $\hat{y}^{a}_{t}=1$ and $\hat{y}^{v}_{t}=1$ for an audible-only event with ground-truth labels $y^{a}_{t}=1$ and $y^{v}_{t}=0$. According to the A metric, this prediction is considered a true positive. However, when both modalities are considered, it becomes evident that the network's prediction is incorrect, as it predicted an audible-visible event instead of an audible-only event. Thus, this should be considered a false positive for the audible-only metric. Consequently, a network that excels at detecting audible-visible events may falsely exhibit high true positive rates for the A and V metrics. This issue becomes particularly prominent when using cross-modal context in a weakly supervised scenario, as methods under these settings inherently tend to detect events as audible-visible. To overcome this, we introduce `audible-only’ and `visible-only’ event proposals, $\hat{y}^{ao}_t$ and $\hat{y}^{vo}_t$, as well as `audible-only’ and `visible-only’ ground-truth labels, ${y}^{ao}_t$ and ${y}^{vo}_t$, that are derived by considering both modalities as \begin{gather}
\small
    \hat{y}^{ao}_{t} = \hat{y}^{a}_{t} \odot (1 - \hat{y}^{v}_{t}),\text{ and } \hat{y}^{vo}_{t} = \hat{y}^{v}_{t} \odot (1 - \hat{y}^{a}_{t}),\\
    {y}^{ao}_{t} = {y}^{a}_{t} \odot (1 - {y}^{v}_{t}),\text{ and }{y}^{vo}_{t,i} = {y}^{v}_{t} \odot (1 - {y}^{a}_{t}).
\end{gather} 
These new event proposals and their corresponding ground-truth labels are used to compute the F-score for audible-only (Ao) and visible-only (Vo) events. Subsequently, we have Type@AVo, which computes the average F-score of Ao, Vo, and AV, and Event@AVo, which computes the overall F-score for all Ao and Vo events within each sample.
 
\section{Experimental Results}
\noindent{\bf Dataset --} LLP \cite{tian2020unified} is the only benchmark dataset for the AVVP task. It comprises 11,849 10-second YouTube video clips spanning 25 event classes, offering a diverse range of video events. Notably, 7,202 of these videos contain multiple event categories. In the LLP dataset, videos are partitioned into training, validation, and testing sets. While the training set includes only video-level annotations, the validation and test sets feature annotations for individual audio and visual events with second-wise temporal boundaries.

\noindent{\bf Implementation Details --} 
 We implement the {\r} branch by employing two self-attention layers, \emph{i.e.}, one for each modality, followed by two fully connected layers that obtain the temporal weights and estimate class probabilities, respectively. For the {\a} branch, we employ CMPAE \cite{gao2023collecting} as the backbone. 
Following previous works \cite{tian2020unified,gao2023collecting}, audio input tokens are extracted through pre-trained VGGish \cite{hershey2017cnn}, and visual tokens are obtained through the pre-trained models ResNet152 \cite{he2016deep} and R(2+1)D \cite{tran2018closer}. The feature dimension for all input tokens, including the class tokens, is set to 512, 
as in
previous approaches. To train {\method}, we utilized the Adam optimizer with an initial learning rate of $5 \times 10^{-4}$ and a batch size of 128 for 15 epochs. The learning rate was decayed by a factor of 0.25 every 6 epochs. Our experiments were performed using PyTorch on an NVIDIA GeForce RTX 3090 GPU.

\subsection{Ablation Studies}
\label{sec: ablations}
In this section, we examine the impact of key components of {\method}, using the Ao, Vo, and AV metrics on LLP.

\noindent{\bf Impact of Event-Aware Contrastive Learning -- } To evaluate the effect of our proposed contrastive loss, $\mathcal{L}^{Anch}_{Evt}$, we compare the performance of our framework against three variants: one without $\mathcal{L}^{Anch}_{Evt}$, and two where it is replaced with the standard NCE loss $\mathcal{L}_{NCE}$ and BCE loss $\mathcal{L}_{BCE}$, the latter using the pseudo-labels employed by $\mathcal{L}^{Anch}_{Evt}$. Table \ref{tab: contrastive loss} demonstrates that standard contrastive learning is ineffective for the AVVP task. While $\mathcal{L}_{BCE}$ yields average improvement, its effectiveness is not consistent across all event types, i.e., it deteriorates performance in detecting the visible-only events in both segment and event levels. Our contrastive loss tailored for this task increases the method’s performance across all event types (see blue numbers in Table~\ref{tab: contrastive loss}). Specifically, it leads to a significant improvement {({\bf{over 2\%}} F-score)} in detecting audible-only events in both segment and event levels. 

\noindent{\bf Impact of Self-Modality-Aware Knowldge -- } In Table~\ref{tab: self-loss}, we ablate our proposed self-modality-aware knowledge distillation loss $\mathcal{L}^{Anch}_{SelfMo}$. The results demonstrate that the additional free modality-aware knowledge obtained by {\a} for {\r} during training eventually enhances the performance of {\a} itself across all event types, specifically resulting in {\bf over a 1\%} improvement in F-score for detecting unaligned audible-visible event types.

\noindent{\bf Impact of Learning Co-occurrence Class Relationships -- } In {\method}, the class tokens and $\mathcal{{L}}^{Anch}_{CoCls}$ are introduced to explicitly model and transfer the cross-class relationships from {\r} to {\a}. In Table~\ref{tab: class relationships}, we conduct ablations on these two components. The results demonstrate that leveraging both components increases the method’s performance {\bf{over a 1\%}} F-score across all event types (see blue numbers in Table~\ref{tab: class relationships}). Additionally, the results reveal that (i) the network can benefit from $\mathcal{{L}}^{Anch}_{CoCls}$ with or without the use of the class tokens and (i) adding the class tokens does not improve the performance on average unless the network leverages $\mathcal{{L}}^{Anch}_{CoCls}$. In these experiments, we also observed that using the class tokens increased the computational complexity of {\r} by 41.7\% Flops (\ie, from 12.7 to 18.0 GFlops) for 25 classes of LLP. However, our novel design in {\method} allows us to benefit from the knowledge learned from them without suffering from their computational overhead in inference. Notably, this novel design can benefit other video understanding tasks where exploiting class dependencies is crucial.

\begin{table}[t]
\scalebox{0.8}
{
  \floatsetup{captionskip=4pt}
  \begin{floatrow}[2]
    \hspace{-20mm}
        \ttabbox%
    {\setlength{\tabcolsep}{1 pt} 
    \renewcommand{\arraystretch}{0.95}
    \begin{tabularx}{0.572\textwidth}{@{}l ccc c ccc @{}}
      \specialrule{.2em}{.1em}{.1em}
     \multirow{2}{*}{Loss}& \multicolumn{3}{c}{Segment-level}&~\phantom & \multicolumn{3}{c}{Event-level}\cr \cmidrule{2-4} \cmidrule{6-8}
      & Ao& Vo& AV&\phantom& Ao& Vo& AV\cr \specialrule{.2em}{.1em}{.1em} 
     $\mathcal{L}_{NCE}$& 46.1& 60.8& 56.3& &40.5&60.6&50.1 \cr 
     $\mathcal{L}_{BCE}$& {48.5}& {61.2} & {57.4} & & {43.5}& {61.1}& {51.2}\cr \specialrule{.2em}{.1em}{.1em} \specialrule{.2em}{.1em}{.1em}
     \multicolumn{1}{c}{\bf-} &  47.3& 61.8& 56.7& &41.6&61.7&50.3 \cr\cmidrule{1-8}
     $\mathcal{L}^{Anch}_{Evt}$& {\bf49.3}& {\bf62.4}& {\bf58.6}& & {\bf44.1}& \bf{62.2}& {\bf52.1} \cr 
     & \blue{(+2.1)}& \blue{(+0.6)}& \blue{(+0.9)}& & \blue{(+2.5)}&\blue{(+0.5)}&\blue{(+0.8)}\cr \specialrule{.2em}{.1em}{.1em}
      \end{tabularx}}
    {\caption{Ablation studies on $\mathcal{L}^{Anch}_{Evt}$.}
  \label{tab: contrastive loss}}
    \ttabbox%
    {
    \setlength{\tabcolsep}{2 pt} 
    \renewcommand{\arraystretch}{1.1}
    
    \begin{tabularx}{0.645\textwidth}{c ccc c ccc}
      \specialrule{.2em}{.1em}{.1em}
      \multirow{2}{*}{$\mathcal{{L}}^{Anch}_{SelfMo}$}& \multicolumn{3}{c}{Segment-level}&\phantom & \multicolumn{3}{c}{Event-level}\cr \cmidrule{2-4}  \cmidrule{6-8}
    &  \multicolumn{1}{c}{Ao}& \multicolumn{1}{c}{Vo}& \multicolumn{1}{c}{AV}&\phantom& \multicolumn{1}{c}{Ao}& \multicolumn{1}{c}{Vo}& \multicolumn{1}{c}{AV}\cr \specialrule{.2em}{.1em}{.1em}
     \xmark &  48.1& 60.9& 57.9& & 42.8&60.6&51.4\cr \cmidrule{1-8}
     \cmark  & {\bf49.3}& {\bf62.4}& {\bf58.6}& & {\bf44.1}& {\bf62.2}& {\bf52.1}\cr 
     & \blue{(+1.2)}& \blue{(+1.5)}& \blue{(+0.7)}& & \blue{(+1.3)}&\blue{(+1.6)}&\blue{(+0.7)}\cr\specialrule{.2em}{.1em}{.1em}
      \end{tabularx}}
    {\caption{Ablation studies on $\mathcal{L}^{Ref}_{SelfMo}$.}
  \label{tab: self-loss}}
  \end{floatrow}}
\end{table}%

\begin{table}[t]
\scalebox{0.80}
{
  \floatsetup{captionskip=4pt}
  \begin{floatrow}[2]
    \hspace{-23mm}
        \ttabbox%
    {\setlength{\tabcolsep}{0.6 pt} 
    \renewcommand{\arraystretch}{0.99}
    \begin{tabularx}{0.64\textwidth}{@{}cc c ccccccc @{}}
      \specialrule{.2em}{.1em}{.1em}
      \multirow{2}{*}{Tkn}& ~~\phantom &\multirow{2}{*}{$\mathcal{{L}}^{Anch}_{CoCls}$} & \multicolumn{3}{c}{Segment-level}&~\phantom & \multicolumn{3}{c}{Event-level}\cr \cmidrule{4-6} \cmidrule{8-10}
      & & & Ao& Vo& AV&\phantom& Ao& Vo& AV\cr \specialrule{.2em}{.1em}{.1em}
      \cmark& & \xmark& 47.5& 60.3& 57.0& &41.9&60.2&50.4 \cr 
     \xmark& &\cmark& 48.2& {62.7}& 57.3& &42.5&{62.5}& 52.0\cr \specialrule{.2em}{.1em}{.1em} \specialrule{.2em}{.1em}{.1em}
     \xmark& &\xmark& 47.8& 60.8& 56.6& &42.2&60.6&49.8 \cr \cmidrule{1-10}
     \cmark& &\cmark& {\bf49.3}&{\bf 62.4}&{\bf58.6}& &{\bf44.1}&{\bf 62.2}& {\bf52.1}\cr &  && \blue{(+1.1)}& \blue{(+1.6)}& \blue{(+2.0)}& & \blue{(+1.9)}&\blue{(+1.6)}&\blue{(+2.3)} \cr\specialrule{.2em}{.1em}{.1em}
      \end{tabularx}}
    {\caption{Ablation studies on class tokens (Tkn) and $\mathcal{{L}}^{Anch}_{CoCls}$.}
  \label{tab: class relationships}}
    \ttabbox%
    {
    \setlength{\tabcolsep}{0.8 pt} 
    \renewcommand{\arraystretch}{0.76}
    
    \begin{tabularx}{0.635\textwidth}{lccccccc}
      \specialrule{.2em}{.1em}{.1em}
      &\multicolumn{3}{c}{Segment-level}&&\multicolumn{3}{c}{Event-level}\cr \cmidrule{2-4} \cmidrule{6-8}
      & Ao& Vo& AV&~\phantom& Ao& Vo& AV\cr \specialrule{.2em}{.1em}{.1em}
      HAN \cite{tian2018audio}& 33.1& 50.7& 48.9& & 31.0&50.1&43.0 \cr \cmidrule{1-8}
     HAN$\circledast$ & {\bf36.3}& {\bf54.1}& {\bf49.4}& & {\bf34.2}& {\bf53.5}& {\bf43.7}\cr
     & \blue{(+3.2)}& \blue{(+3.4)}& \blue{(+0.5)}& & \blue{(+3.2)}& \blue{(+3.4)}& \blue{(+0.7)}\cr \specialrule{.2em}{.1em}{.1em} \specialrule{.2em}{.1em}{.1em}
    CMPAE \cite{gao2023collecting}& 48.2& 57.9& 57.5& & 43.6&57.5&49.6 \cr \cmidrule{1-8}
            CMPAE$\circledast$& {\bf49.3}& {\bf62.4}& {\bf58.6}& & {\bf44.1}& {\bf62.2}& {\bf52.1}\cr 
    & \blue{(+1.1)}& \blue{(+4.5)}& \blue{(+1.1)}& & \blue{(+0.5)}& \blue{(+4.7)}& \blue{(+2.5)}\cr \specialrule{.2em}{.1em}{.1em}
      \end{tabularx}}
    {\caption{Framework generalization on LLP. $\circledast$ indicates that the network is embedded in {\method} as the {\a} branch.}
  \label{tab: framework generalization}}
  \end{floatrow}}
\end{table}%

\noindent{\bf Framework Generalization --} {\method} is a general learning framework, which means that any existing AVVP method can be embedded in our framework to (i) explicitly optimize the integration of cross-modal information and (ii) benefit from modeling co-occurrence class relationships in an end-to-end manner. In Table \ref{tab: framework generalization}, we present two examples: HAN \cite{tian2020unified}, a common architecture for AVVP, and CMPAE \cite{gao2023collecting}. The results demonstrate that our proposed framework significantly enhances the performance of HAN and CMPAE, \ie, on average, the F-score is increased by {\bf{2.4\%}} in both networks. This improvement is achieved in an end-to-end manner without requiring additional annotations or altering their network architecture.

\subsection{State-of-the-Art Comparison} 
In this section, we compare the performance of {\method} with current state-of-the-art approaches using both traditional and our proposed metrics. The results of the previous methods under the new metrics were reproduced using their available code whenever accessible. Note that to obtain the event proposals from the estimated output probabilities, CMPAE \cite{gao2023collecting} utilizes a set of `selected thresholds’ for different event categories of LLP, while in the standard setting used by other state-of-the-art approaches, a fixed threshold is applied to all event classes. For a fair comparison, we also present the CMPAE’s results under the standard setting.

Table~\ref{tab: SOTA-weak} presents comparative results on the LLP dataset when only video-level labels are used for training. The results demonstrate that our proposed approach {\method} achieves new state-of-the-art results on all new metrics, {as well as AV under both standard and `selected thresholds’ settings}. {\method} exhibits over {\bf 1.1\%} F-score improvement on most new metrics. Specifically, it significantly increases Vo and Event@Avo by {\bf 3.6\%} and {\bf 2.0\%} F-score at the segment level, and Vo, AV, and Type@Avo by {\bf 2.2\%}, {\bf 2.7\%}, and {\bf 2.4\%} F-score at the event level (under the standard setting), {and Type@Avo by {\bf 2.1\%} F-score at the event level (under the `selected thresholds’ setting).} On average across new metrics and AV, our proposed method improves the state-of-the-art results by 1.9\% and {1.2\% F-score} under standard and selected threshold settings, respectively. 


\noindent{\bf Analysis of Metrics --} In Section~\ref{sec: metrics}, we stated that the traditional A and V metrics do not adequately reflect the method’s ability to detect audible-only and visible-only events, respectively. As in the A and V metrics, apart from counting audible-only and visible-only events, respectively, they mistakenly include audible-visible events as true positive. This can have a detrimental impact on the analysis of the method's performance. For example, using the traditional A metric, HAN and MA \cite{wu2021exploring} appear to show nearly identical performance in detecting audible-only events at 60.1\% and 60.3\% F-score (at segment-level), respectively. However, when we evaluate them using our proposed Ao metric that counts correctly audible-only events, we observe that MA outperforms HAN with a margin of 10.4\%.  Similarly, based on the V metric (at event-level), CMPAE \cite{gao2023collecting} outperforms JoMoLD \cite{cheng2022joint} in detecting visible-only events, while our proposed Vo metric reveals that JoMoLD achieves a better performance than CMPAE. Consequently, the traditional Type@AV and Event@AV have similar issues.

\begin{table}[t]
\caption{Weakly supervised AVVP results on the LLP dataset when utilizing video-level labels for training. All methods use VGGish, ResNet152, and R(2+1)D to generate the input tokens. `T@' and `E@' refers `Type@AV' and `Event@AV', respectively. {$\dag$} indicates the results are under {`selected thresholds'} setting. The best and the second-best results are in {\bf Bold} and \underline{underlined}, respectively. \vspace{-3mm}}
  \label{tab: SOTA-weak}
\centering
\scalebox{0.75}
{ \hspace{-7.8mm}
  \setlength{\tabcolsep}{0.1pt} 
    \renewcommand{\arraystretch}{0.85} 
  \begin{tabular}{@{}l c ccccccccc c ccccccccc @{}}\specialrule{.2em}{.1em}{.1em}
    \multirow{2}{*}{Method}&\phantom & \multicolumn{9}{c}{Segment-level (\%)}&\phantom & \multicolumn{9}{c}{Event-level (\%)}\cr \cmidrule{3-11} \cmidrule{13-21}
    & \phantom & A&Ao& V&Vo& AV& T@&T@o & E@&E@o&\phantom& A&Ao& V&Vo& AV& T@&T@o & E@&E@o\cr \specialrule{.2em}{.1em}{.1em}
    AVE \cite{tian2018audio}&  & 47.2&--& 37.1&--& 35.4& 39.9&~~--~ &41.6&--&~~& 40.4&--& 34.7&--&31.6&35.5&~~--~& 36.5&--\cr
    AVSDN \cite{lin2019dual} & & 47.8&--&52.0&-- & 37.1& 45.7&--& 50.8&--&~~& 34.1&--& 46.3&--& 26.5&35.6&--& 37.7&--\cr
    CVCMS \cite{lin2021exploring}& & 59.2&--& 59.9&--& 53.4& 57.5&--& 58.1&--&~~& 51.3&--& 55.5&--& 46.2& 51.0&--& 49.7&-- \cr
    
    MM-Pyr \cite{yu2022mm} & & 60.9&--& 54.4&--& 50.0& 55.1&--&57.6&--& & 52.7&--& 51.8&--& 44.4& 49.9&--& 50.5&--\cr
    DHHN \cite{jiang2022dhhn} & & 61.3&--& 58.3&--& 52.9& 57.5&--&58.1&--& & 54.0&--&55.1&-- & 47.3& 51.5&--& 51.5&--\cr
    MGN \cite{mo2022multi} & & 60.8&--& 55.4&--& 50.0& 55.5&--&57.2&--& & 51.1&--& 52.4&--& 44.4& 49.3&--& 49.1&--\cr \cmidrule{3-21}
    HAN \cite{tian2020unified}& &{60.1}&33.1&52.9&50.7&48.9&54.0&44.3&55.4&24.1& &51.3&31.0&48.9&50.1&43.0&47.7&41.4&48.0&22.0\cr
    MA \cite{wu2021exploring}& & {60.3}&43.5& 60.0&53.6& 55.1& 58.9&50.8&57.9&32.0& & 53.6&38.8& 56.4&52.4& 49.0& 53.0&46.8& 50.6&27.8\cr
    JoMoLD \cite{cheng2022joint}& & {61.3}&46.2&\underline{63.8}&\underline{58.8} &{57.2} &60.8&\underline{54.4} &59.9&35.4 &~~& 53.9&40.9&59.9&\underline{59.0} &\underline{49.6} & 54.5&\underline{50.3}& 52.5&31.1\cr
    CMPAE \cite{gao2023collecting}& &\bf{{63.9}}&\underline{48.2} & 63.3&57.9& \underline{57.5}& \underline{61.2}&54.2&\bf{62.1}&\underline{35.7}& & \bf{57.1}&\underline{43.6}& \underline{60.6}&57.5& \underline{49.6}& \underline{55.8}&\underline{50.3}& \underline{55.6}&\underline{32.0}\cr \cmidrule{1-21} 
    {{\method}} & & {\underline{63.5}}&{\bf49.3}& {\bf64.5}&{\bf62.4}& {\bf58.6}& {\bf62.1}&{\bf56.7}&\underline{61.8}&{\bf37.7}& & \underline{56.2}&{\bf44.1}& {\bf62.0}&{\bf62.2}& {\bf52.1}&{\bf56.7}&{\bf52.7} & {\bf54.5}&{\bf33.4}\cr 
    & & &\blue{(+1.1)}& &\blue{(+3.6)}& \blue{(+1.1)}& &\blue{(+1.4)}&~~&\blue{(+2.0)}& & &\blue{(+0.5)}& &\blue{(+2.2)}& \blue{(+2.7)}& &\blue{(+2.4)}& &\blue{(+1.4)}\cr\specialrule{.2em}{.1em}{.1em} \specialrule{.2em}{.1em}{.1em}
    CMPAE{$\dag$} \cite{gao2023collecting}& &{\bf64.2}&48.9 & 66.4&61.7& 59.2& 63.3&55.5&{\bf62.8}&37.6& & 56.6&43.8& 63.7&61.1& 51.8& 51.9&50.3& {\bf55.7}&34.0\cr \cmidrule{1-21}
    {{\method}}{$\dag$} & & {\bf 64.2}&{\bf49.6}& \bf{67.1}&{\bf63.1}& \bf{59.8}& \bf{63.8}&{\bf57.0}&{61.9}&{\bf38.4}& & {\bf57.1}&{\bf45.0}& {\bf64.8}&{\bf62.7}& {\bf52.8}&{\bf58.2}&{\bf52.4} & {55.5}&{\bf34.6}\cr 
    & & &\blue{(+0.7)}& &\blue{(+1.4)}& \blue{(+0.6)}& &\blue{(+1.5)}&~~&\blue{(+0.8)}& & &\blue{(+1.2)}& &\blue{(+1.6)}& \blue{(+1.0)}& &\blue{(+2.1)}& &\blue{(+0.6)}\cr\specialrule{.2em}{.1em}{.1em}
  \end{tabular}
  }\vspace{2mm}
\end{table}

\begin{table}[t]

\centering
\caption{Weakly supervised AVVP results on the LLP dataset when utilizing video-level labels and CLIP and CLAP annotations for training. All methods use VGGish, ResNet152, and R(2+1)D to generate the input tokens (see Table~\ref{tab: SOTA-weak} for more definitions and terms).\vspace{-3mm}}
\scalebox{0.75}
{ \hspace{-7.3mm}
  \setlength{\tabcolsep}{0.02pt} 
    \renewcommand{\arraystretch}{1.0} 
  \begin{tabular}{@{}l c ccccccccc c ccccccccc @{}}\specialrule{.2em}{.1em}{.1em}
    \multirow{2}{*}{Method}&\phantom & \multicolumn{9}{c}{Segment-level (\%)}&\phantom & \multicolumn{9}{c}{Event-level (\%)}\\ \cmidrule{3-11} \cmidrule{13-21}
    & \phantom & A&Ao& V&Vo& AV& T@&T@o & E@&E@o&\phantom& A&Ao& V&Vo& AV& T@&T@o & E@&E@o\\ \specialrule{.2em}{.1em}{.1em}
    JoMoLD \cite{cheng2022joint}& & 61.5&46.3& \underline{64.7}&64.8& 58.3& 61.5&56.5&60.3&38.0& & 54.7&41.5& 61.5&63.5& 52.4& 56.2&52.5&53.1&33.6\\
    CMPAE \cite{gao2023collecting}& & \underline{ 64.1}&48.5&64.3&\underline{66.9} & \underline{58.9}& \underline{62.4}&\underline{58.1}&\underline{62.0}&{40.0}& & \underline{57.0}&42.7& 62.1&\underline{66.0}& \underline{53.5}& \underline{57.6}&\underline{54.1}& 55.0&35.1\\ 
    VALOR \cite{lai2023modality} & & 61.8&\underline{49.0}& {\bf65.9}&66.7& 58.4&62.0&57.9 &61.5&\underline{40.8}& &55.4&\underline{44.2} &\underline{62.6}&65.0 &52.2 &56.7&53.8 & \underline{54.2}&\underline{35.8}\\\cmidrule{1-21}
    {{\method}} & & \bf{64.2}&{\bf{51.2}}& {64.4}&{\bf67.7}& {\bf59.3}& {\bf62.6}&{\bf59.4}&{\bf62.5}&{\bf41.3}&~~& \bf{57.6}&{\bf45.5} &{\bf63.2}&{\bf67.0} & {\bf54.2}&{\bf57.9}&{\bf55.6} & \bf{55.6}&{\bf36.4}\cr
    & & &\blue{(+2.2)}& &\blue{(+0.8)}& \blue{(+0.4)}& &\blue{(+1.3)}&&\blue{(+0.5)}& & &\blue{(+1.3)}& &\blue{(+1.0)}& \blue{(+0.7)}& &\blue{(+1.5)}& &\blue{(+0.6)}\cr\specialrule{.2em}{.1em}{.1em} 
  \end{tabular}
  }
  
  \label{tab: clip and clap}
\end{table}

\begin{figure}[t]
\CenterFloatBoxes
\begin{floatrow}
\ffigbox
  {\includegraphics[width=1.0\linewidth]{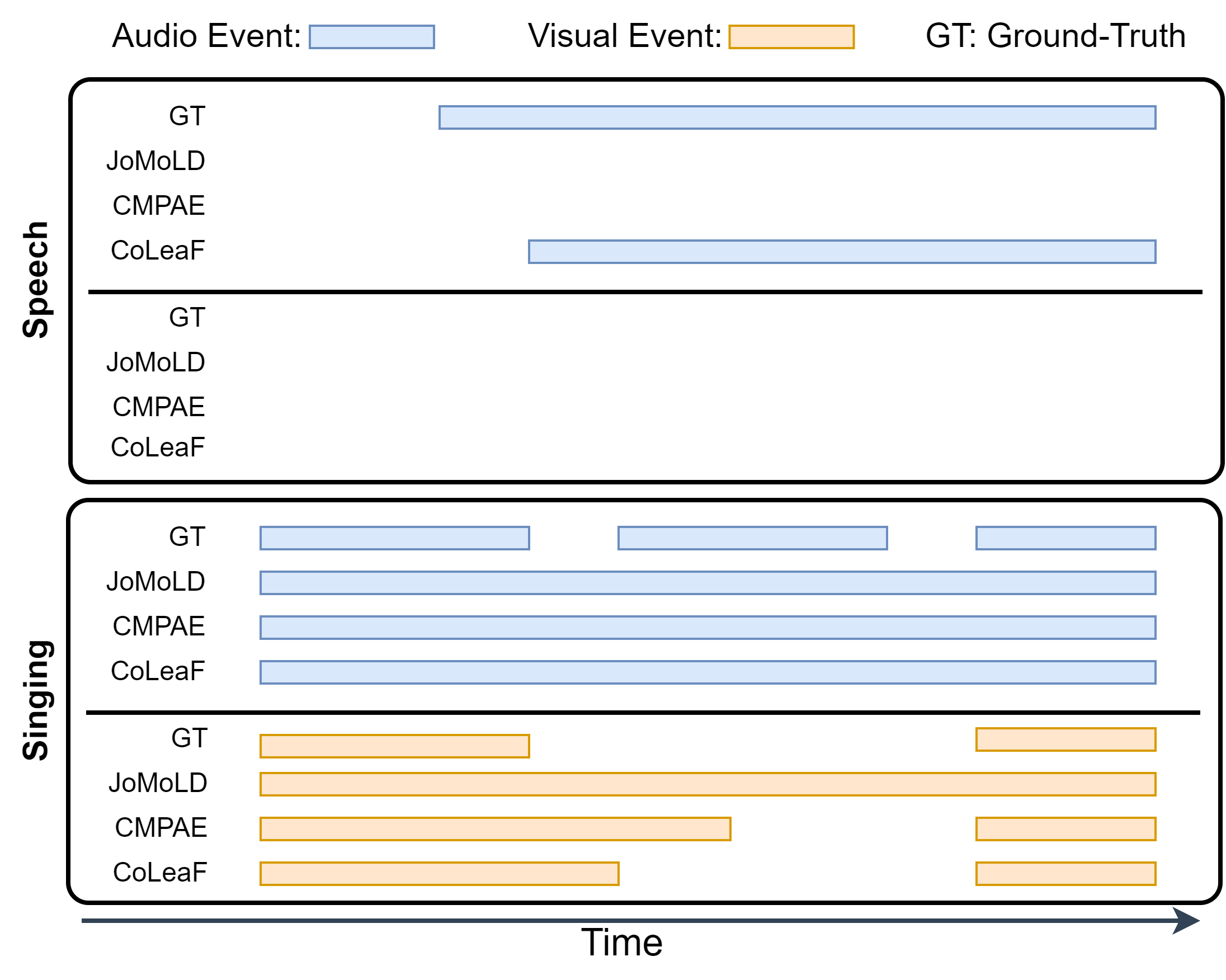}
  \caption{Qualitative comparison to previous AVVP approaches (JoMoLD \cite{cheng2022joint} and CMPAE \cite{gao2023collecting}) training with weak labels.}
  \label{fig:qualitative}}
\killfloatstyle
\ttabbox
{ \scalebox{0.9}
{
\begin{tabular}{@{}lcc @{}}\specialrule{.2em}{.1em}{.1em}
    {Method} & {AV (Seg)}& {AV (Evn)}\cr \specialrule{.2em}{.1em}{.1em}
     HAN \cite{tian2020unified}& 35.0& 41.4\cr
     MA \cite{wu2021exploring} & 37.9& \underline{44.8}\cr
     JoMoLD \cite{cheng2022joint} & 36.4& 41.2\cr
     CMPAE \cite{gao2023collecting} & \underline{39.7}& 43.8\cr \cmidrule{1-3}
     {\method}& {\bf41.5}~\blue{(+1.8)}& {\bf47.8}~\blue{(+3.0)}\cr\specialrule{.2em}{.1em}{.1em} 
  \end{tabular}
  }}
  {\caption{Weakly supervised DAVE results on the UnAV-100 dataset. All methods use pre-trained VGGish \cite{hershey2017cnn} and two-stream I3D \cite{ carreira2017quo} to generate the input tokens. The best and the second-best results are in {\bf Bold} and \underline{underlined}, respectively.}
  \label{tab: AVE results}}
\end{floatrow}
\end{figure}

\noindent{\bf Qualitative Results --} 
In Fig.~\ref{fig:qualitative}, we qualitatively compare {\method} with the state-of-the-art JoMoLD \cite{cheng2022joint} and CMPAE \cite{gao2023collecting} on a test video sample of LLP. For the {`Speech'} event, while JoMoLD and CMPAE cannot detect it, {\method} recognizes it correctly as audible-only. While all methods detect the audible-visible event {`Singing'}, {\method}'s event predictions have a better overlap with the ground-truth annotations. 

\noindent{\bf CLIP and CLAP Annotations --} Table~\ref{tab: SOTA-weak} addresses
the weakly supervised AVVP task without the need for additional annotations.
The authors in \cite{lai2023modality} generate offline dense modality-specific annotations for the LLP dataset
with
large-scale pretrained models, CLIP \cite{radford2021learning}, and CLAP \cite{wu2023large}. In Table~\ref{tab: clip and clap}, we present comparative results with recent state-of-the-art approaches in which the networks are additionally trained on these annotations. The results of JoMoLD \cite{cheng2022joint} and CMPAE \cite{gao2023collecting} were obtained by adapting their available code. The results demonstrate that in this scenario with the modality-specific pseudo labels, {\method} still improves the state-of-the-art approaches consistently across all new metrics and AV F-score.

    

\subsection{Generalization on the Large-Scale UnAV-100 Dataset}
To demonstrate the superior generalization of our proposed framework on a different, yet challenging dataset, we present its results to the large-scale UnAV-100 dataset, a benchmark for the Dense Audio-Visual Event Localization (DAVE) task \cite{geng2023dense}. UnAV-100 includes videos with multiple events and encompasses a diverse range of 100 distinct audio-visible action types, four times more than the LLP dataset. Furthermore, UnAV-100 features untrimmed videos ranging from 0.2 to 60 seconds, significantly increasing the dataset's complexity. To obtain weakly supervised results, we derived video-level labels from the videos' temporal annotations (see supp. file for the implementation details for this task). The results of other state-of-the-art approaches were provided through their available code. Comparative results on UnAV-100 using AV F-score are available in Table \ref{tab: AVE results}. It is shown that {\method} outperforms the other approaches by \textbf{1.8\%} and \textbf{3.0\%} AV F-score at the segment and event levels, respectively, and achieves state-of-the-art results on this dataset.

\section{Conclusion}
We introduce {\method}, a novel learning framework designed to: (i) explicitly optimize integrating cross-modal contexts in the embedding space for the weakly supervised AVVP task; and (ii) benefit from explicitly modelling cross-class relationships during training without adding computational overhead at inference. Having introduced novel metrics to better understand an AVVP method's capability in detecting audible-only and visible-only events, we evaluate {\method} on LLP, the only benchmark dataset for the weakly supervised AVVP task, and UnAV-100, a challenging benchmark dataset for the weakly supervised DAVE task. Our method achieves new state-of-the-art results on both datasets. Ablation studies demonstrate the effectiveness of {\method}'s key components. Future work will adapt our end-to-end framework to benefit from language models.

\section*{Acknowledgement}
This research is supported by UKRI EPSRC Platform Grant EP/P022529/1, and EPSRC BBC Prosperity Partnership AI4ME: Future Personalised Object-Based Media Experiences Delivered at Scale Anywhere EP/V038087/1.

\bibliographystyle{splncs04}
\bibliography{main}

\begin{thebibliography}{10}
\providecommand{\url}[1]{\texttt{#1}}
\providecommand{\urlprefix}{URL }
\providecommand{\doi}[1]{https://doi.org/#1}

\bibitem{carreira2017quo}
Carreira, J., Zisserman, A.: {Quo Vadis, Action Recognition? A New Model and the Kinetics Dataset}. In: Proceedings of the IEEE/CVF Conference on Computer Vision and Pattern Recognition. pp. 6299--6308 (2017)

\bibitem{cheng2022joint}
Cheng, H., Liu, Z., Zhou, H., Qian, C., Wu, W., Wang, L.: {Joint-Modal Label Denoising for Weakly-Supervised Audio-Visual Video Parsing}. In: European Conference on Computer Vision. pp. 431--448. Springer (2022)

\bibitem{gao2023collecting}
Gao, J., Chen, M., Xu, C.: {Collecting Cross-Modal Presence-Absence Evidence for Weakly-Supervised Audio-Visual Event Perception}. In: Proceedings of the IEEE/CVF Conference on Computer Vision and Pattern Recognition. pp. 18827--18836 (2023)

\bibitem{geng2023dense}
Geng, T., Wang, T., Duan, J., Cong, R., Zheng, F.: {Dense-Localizing Audio-Visual Events in Untrimmed Videos: A Large-Scale Benchmark and Baseline}. In: Proceedings of the IEEE/CVF Conference on Computer Vision and Pattern Recognition. pp. 22942--22951 (2023)

\bibitem{gutmann2010noise}
Gutmann, M., Hyv{\"a}rinen, A.: {Noise-Contrastive Estimation: A New Estimation Principle for Unnormalized Statistical Models}. In: Proceedings of the 13th International Conference on Artificial Intelligence and Statistics. pp. 297--304. JMLR Workshop and Conference Proceedings (2010)

\bibitem{he2016deep}
He, K., Zhang, X., Ren, S., Sun, J.: {Deep Residual Learning for Image Recognition}. In: Proceedings of the IEEE/CVF Conference on Computer Vision and Pattern Recognition. pp. 770--778 (2016)

\bibitem{hershey2017cnn}
Hershey, S., Chaudhuri, S., Ellis, D.P., Gemmeke, J.F., Jansen, A., Moore, R.C., Plakal, M., Platt, D., Saurous, R.A., Seybold, B., et~al.: {CNN Architectures for Large-Scale Audio Classification}. In: IEEE International Conference on Acoustics, Speech and Signal Processing. pp. 131--135. IEEE (2017)

\bibitem{jiang2022dhhn}
Jiang, X., Xu, X., Chen, Z., Zhang, J., Song, J., Shen, F., Lu, H., Shen, H.T.: {DHHN: Dual Hierarchical Hybrid Network for Weakly-Supervised Audio-Visual Video Parsing}. In: Proceedings of the 30th ACM International Conference on Multimedia. pp. 719--727 (2022)

\bibitem{lai2023modality}
Lai, Y.H., Chen, Y.C., Wang, Y.C.F.: {Modality-Independent Teachers Meet Weakly-Supervised Audio-Visual Event Parser}. {Advances in Neural Information Processing systems}  (2023)

\bibitem{li2022learning}
Li, G., Wei, Y., Tian, Y., Xu, C., Wen, J.R., Hu, D.: {Learning to Answer Questions in Dynamic Audio-Visual Scenarios}. In: Proceedings of the IEEE/CVF Conference on Computer Vision and Pattern Recognition. pp. 19108--19118 (2022)

\bibitem{lin2019dual}
Lin, Y.B., Li, Y.J., Wang, Y.C.F.: {Dual-Modality Seq2Seq Network for Audio-Visual Event Localization}. In: IEEE International Conference on Acoustics, Speech and Signal Processing. pp. 2002--2006. IEEE (2019)

\bibitem{lin2021exploring}
Lin, Y.B., Tseng, H.Y., Lee, H.Y., Lin, Y.Y., Yang, M.H.: {Exploring Cross-Video and Cross-Modality Signals for Weakly-Supervised Audio-Visual Video Parsing}. Advances in Neural Information Processing Systems  \textbf{34},  11449--11461 (2021)

\bibitem{mo2022multi}
Mo, S., Tian, Y.: {Multi-Modal Grouping Network for Weakly-Supervised Audio-Visual Video Parsing}. Advances in Neural Information Processing Systems  \textbf{35},  34722--34733 (2022)

\bibitem{nadeem2023cad}
Nadeem, A., Hilton, A., Dawes, R., Thomas, G., Mustafa, A.: {CAD--Contextual Multi-modal Alignment for Dynamic AVQA}. Proceedings of the IEEE/CVF Winter Conference on Applications of Computer Vision  (2023)

\bibitem{radford2021learning}
Radford, A., Kim, J.W., Hallacy, C., Ramesh, A., Goh, G., Agarwal, S., Sastry, G., Askell, A., Mishkin, P., Clark, J., et~al.: { Quantify Classificationl Models from Natural Language Supervision}. In: International Conference on Machine Learning. pp. 8748--8763 (2021)

\bibitem{sensoy2018evidential}
Sensoy, M., Kaplan, L., Kandemir, M.: {Evidential Deep Learning to Quantify Classification Uncertainty}. Advances in Neural Information Processing Systems  \textbf{31} (2018)

\bibitem{tian2020unified}
Tian, Y., Li, D., Xu, C.: {Unified Multisensory Perception: Weakly-Supervised Audio-Visual Video Parsing}. In: European Conference in Computer Vision. pp. 436--454. Springer (2020)

\bibitem{tian2018audio}
Tian, Y., Shi, J., Li, B., Duan, Z., Xu, C.: {Audio-Visual Event Localization In Unconstrained Videos}. In: European Conference on Computer Vision. pp. 247--263 (2018)

\bibitem{tran2018closer}
Tran, D., Wang, H., Torresani, L., Ray, J., LeCun, Y., Paluri, M.: {A Closer Look at Spatiotemporal Convolutions for Action Recognition}. In: Proceedings of the IEEE/CVF Conference on Computer Vision and Pattern Recognition. pp. 6450--6459 (2018)

\bibitem{vaswani2017attention}
Vaswani, A., Shazeer, N., Parmar, N., Uszkoreit, J., Jones, L., Gomez, A.N., Kaiser, {\L}., Polosukhin, I.: {Attention Is All You Need}. Advances in Neural Information Processing systems  \textbf{30} (2017)

\bibitem{wu2021exploring}
Wu, Y., Yang, Y.: {Exploring Heterogeneous Clues for Weakly-Supervised Audio-Visual Video Parsing}. In: Proceedings of the IEEE/CVF Conference on Computer Vision and Pattern Recognition. pp. 1326--1335 (2021)

\bibitem{wu2023large}
Wu, Y., Chen, K., Zhang, T., Hui, Y., Berg-Kirkpatrick, T., Dubnov, S.: {Large-Scale Contrastive Language-Audio Pretraining with Feature Fusion and Keyword-to-Caption Augmentation}. In: IEEE International Conference on Acoustics, Speech and Signal Processing. pp.~1--5 (2023)

\bibitem{Xia_2022_CVPR}
Xia, Y., Zhao, Z.: {Cross-Modal Background Suppression for Audio-Visual Event Localization}. In: Proceedings of the IEEE/CVF Conference on Computer Vision and Pattern Recognition. pp. 19989--19998 (June 2022)

\bibitem{yu2022mm}
Yu, J., Cheng, Y., Zhao, R.W., Feng, R., Zhang, Y.: {Mm-Pyramid: Multimodal Pyramid Attentional Network for Audio-Visual Event Localization and Video Parsing}. In: Proceedings of the 30th ACM International Conference on Multimedia. pp. 6241--6249 (2022)

\bibitem{Yun_2021_ICCV}
Yun, H., Yu, Y., Yang, W., Lee, K., Kim, G.: {Pano-AVQA: Grounded Audio-Visual Question Answering on 360deg Videos}. In: Proceedings of the IEEE/CVF International Conference on Computer Vision. pp. 2031--2041 (October 2021)

\bibitem{zhou2023improving}
Zhou, J., Guo, D., Zhong, Y., Wang, M.: {Improving Audio-Visual Video Parsing with Pseudo Visual Labels}. arXiv preprint arXiv:2303.02344  (2023)

\end{thebibliography}
\end{document}